\documentclass{article}

\usepackage{arxiv}

\usepackage[utf8]{inputenc} 
\usepackage[T1]{fontenc}    
\usepackage{hyperref}       
\usepackage{url}            
\usepackage{booktabs}       
\usepackage{amsmath,amsfonts,amssymb,amsthm}
\usepackage{nicefrac}       
\usepackage{microtype}      
\usepackage{lipsum}		
\usepackage{graphicx}
\usepackage{natbib}
\usepackage{doi}
\usepackage{tablefootnote}
\usepackage{relsize}
\usepackage{multicol}
\usepackage{optidef} 
\usepackage{multirow}
\usepackage{makecell}
\usepackage{floatrow}
\newfloatcommand{capbtabbox}{table}[][\FBwidth]
\allowdisplaybreaks

\title{Feed-Forward Neural Networks as a Mixed-Integer Program}

\author{ \href{https://orcid.org/0000-0002-2063-1622}{\includegraphics[scale=0.06]{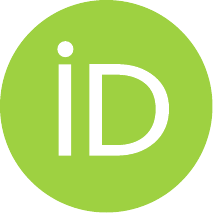}\hspace{1mm}Navid Aftabi} \\
	Industrial Engineering Department \\
	Clemson University\\
	Clemson, SC 29631 \\
	\texttt{naftabi@clemson.edu} \\
	\And
	Nima Moradi \\
	Information Systems Engineering\\
	Concordia University\\
    Montreal, QC H3G 1M8, Canada \\
	\texttt{nima.moradi@mail.concordia.ca} \\
	\AND
	Fatemeh Mahroo \\
	Industrial Engineering Department \\
	Clemson University\\
	Clemson, SC 29631 \\
	\texttt{fmahroo@clemson.edu} \\
}

\renewcommand{\shorttitle}{Feedforward-NNs as MIP}

\hypersetup{
pdftitle={Feedforward-NNs as MIP},
pdfsubject={CS-OR},
pdfauthor={Navid Aftabi, Nima Moradi, Fatemeh Mahroo},
pdfkeywords={Mixed-integer programming, Deep learning, Neural networks, Optimization, Operations research},
}

\begin{document}
\maketitle

\begin{abstract}
	Deep neural networks (DNNs) are widely studied in various applications. A DNN consists of layers of neurons that compute affine combinations, apply nonlinear operations, and produce corresponding activations. The rectified linear unit (ReLU) is a typical nonlinear operator, outputting the max of its input and zero. In scenarios like max pooling, where multiple input values are involved, a fixed-parameter DNN can be modeled as a mixed-integer program (MIP). This formulation, with continuous variables representing unit outputs and binary variables for ReLU activation, finds applications across diverse domains. This study explores the formulation of trained ReLU neurons as MIP and applies MIP models for training neural networks (NNs). Specifically, it investigates interactions between MIP techniques and various NN architectures, including binary DNNs (employing step activation functions) and binarized DNNs (with weights and activations limited to $-1,0,+1$). The research focuses on training and evaluating proposed approaches through experiments on handwritten digit classification models. The comparative study assesses the performance of trained ReLU NNs, shedding light on the effectiveness of MIP formulations in enhancing training processes for NNs.
\end{abstract}


\section{Introduction}
The advancement of digital computing has led to the creation of intelligent machines capable of emulating cognitive functions. These machines learn from previous experiences and tackle complex issues in situations that differ from past occurrences, a field known as machine learning (ML). ML has garnered significant interest among researchers. It focuses on developing a function to produce the most accurate output for a given input while minimizing errors \citep{jordan2015machine}. ML models mainly fall into four categories: supervised, unsupervised, semi-supervised, and reinforcement learning \citep{agarwal2023reinforcement}. 

Supervised learning in ML involves learning a function that closely approximates the relationship between given input and output values (known as ground-truth values) within an acceptable margin of error. Conversely, unsupervised learning does not have response variables available, and its primary aim is to uncover the inherent characteristics of the observations \citep{el2015machine}. ML models' core principles and successes are largely due to interdisciplinary research at the junction of computer science, statistics, and Operations Research (OR) \citep{bennett2006interplay}. \cite{gambella2020optimization} suggests that the relationship between ML and OR can be examined from three perspectives: (1) the application of ML to management science issues, (2) the use of ML in solving optimization problems, and (3) the formulation of ML problems as optimization challenges. 

Most ML models, including those for regression, classification, clustering, deep learning (DL), machine teaching, and empirical model learning, can be formulated as optimization problems \citep{gambella2020optimization}. These models often aim to optimize objectives such as training error, fit measurement, and cross-entropy. Efficient solutions to these optimization problems are crucial for the reliable training and subsequent adoption of ML models. Each model has unique features that need specific consideration, especially in minimizing the corresponding loss function effectively and reliably, a critical aspect in most learning models. Therefore, this paper focuses primarily on the third aspect of the interaction between ML and OR, which involves formulating ML problems as optimization problems. Given the broad application and success of DL in areas like computer vision, natural language processing, and autonomous driving, this study specifically concentrates on feedforward deep neural networks (FF-DNNs). 

Recent surveys on the interplay between optimization, ML, and OR have highlighted the importance of the role of each community in empowering others. \cite{bottou2018optimization} and \cite{curtis2017optimization} share a common focus on applying optimization techniques in ML. They explore how various optimization methods, particularly those involving stochastic gradient techniques, are integral to solving ML problems; \cite{bengio2021machine} provides a comprehensive overview of how ML techniques are utilized to solve combinatorial optimization problems; \cite{gambella2020optimization} explores the intersection of optimization, OR, and ML. It surveys various ML approaches within an OR framework highlighting the role of numerical optimization techniques in these areas; \cite{liu2021algorithms} provides a comprehensive review of methods for verifying the properties of DNNs. It focuses on approaches, including optimization and OR, that ensure a network behaves as expected across its entire input space, addressing the need for formal verification in applications where incorrect outputs could have serious consequences.

MIP formulation of FF-DNNs is discussed in \cite{gambella2020optimization} and \cite{liu2021algorithms}. However, due to the aim of the study, the various ways of formulating FF-DNNs as the MIP are not discussed rigorously. This paper presents MIP models both for trained and training FF-DNNs in detail. This paper also emphasizes the strengths and limitations of these formulations from a mathematical optimization viewpoint by conducting a comparative experiment. This aims to encourage further development in mathematical programming within FF-DNNs. The contributions of the present work are listed as follows:

\begin{itemize}
    \item We rigorously study the various MIP models and valid inequalities proposed for formulating the FF-DNNs as a mathematical program.
    \item We conduct a comparative experimental analysis and highlight the strengths and limitations of each approach from a mathematical optimization viewpoint.
    \item Our work adds to the existing research on the synergy between optimization, ML, and OR. It offers insights to promote further advancements in mathematical programming, particularly in FF-DNNs.
\end{itemize}

Following this section, the background and reviews of the related works in the literature are stated in Section \ref{background}. Section \ref{FF-DNN} presents the MIP formulations of trained and training FF-DNNs. Section \ref{experiments} provides the computational experiment to compare the MIP formulations. Finally, Section \ref{conclusions} states the discussions, conclusions, and suggestions for future studies.

\section{Background and related works}\label{background}
This section covers the concept and background of artificial neural networks, DL models, and various architectures of deep neural networks. We also discuss the learning procedure of deep neural networks using backpropagation. 
\subsection{Artificial Neural Networks}

Dr. Robert Hecht-Nielsen, credited with developing one of the earliest neurocomputers, provided a simple yet straightforward definition of a neural network (NN), specifically an artificial neural network (ANN). According to him, an NN is a computational system comprising many simple, "highly interconnected processing elements." These elements process information by dynamically responding to external inputs \citep{maureen1989neural}. Initially, NNs were primarily driven by the ambition to mimic biological neural systems. Over time, their development has transitioned towards more engineering-focused goals, emphasizing achieving successful outcomes in specific tasks. The NNs are composed of different components \citep{goodfellow2016deep}:

\begin{itemize}
    \item \textbf{Neurons (Perceptrons):} NNs consist of artificial neurons inspired by the structure of biological neurons. Each artificial neuron takes in inputs and generates a singular output, which can subsequently be transmitted to numerous other neurons.
    \item \textbf{Connections and weights (Synaptic Weights):} The NN comprises numerous connections, where the output of one neuron serves as the input to another. Each connection in the network has an assigned weight, signifying its relative importance. A single neuron within this network may have multiple input and output connections, and its unique weight characterizes each of these connections.
    \item \textbf{Propagation (Activation) Function:} In the NN, the propagation function calculates a neuron's input by summing the weighted outputs from its preceding neurons and their connections. Furthermore, this function might incorporate a bias term added to the weighted sum's outcome.
\end{itemize}

\subsubsection{Neurons of ANNs}

\begin{figure}[!t]
    \centering
    \includegraphics[width=0.6\linewidth]{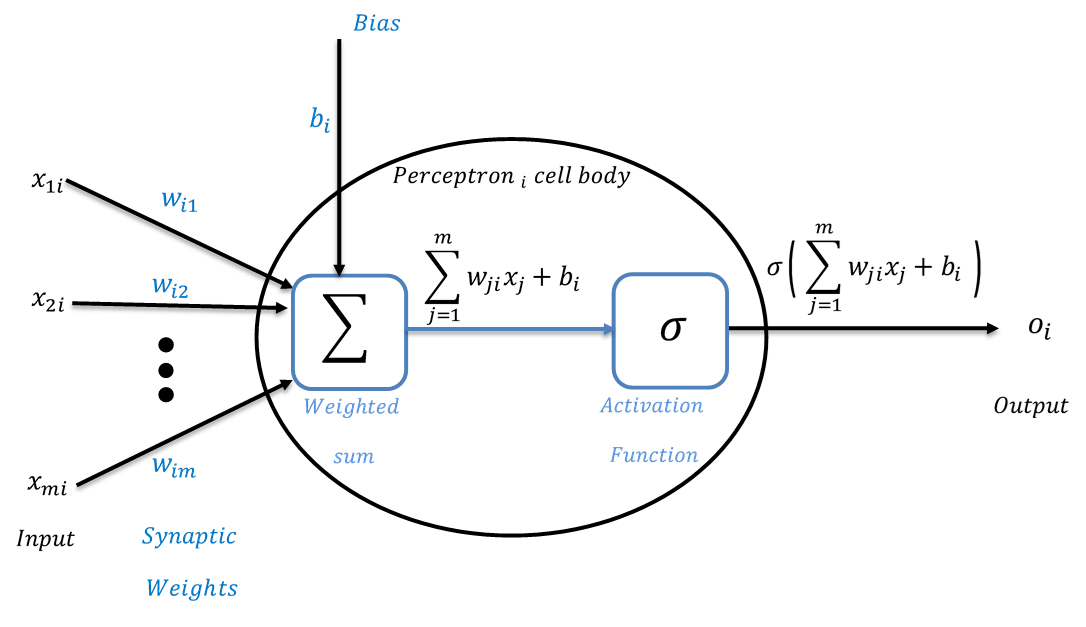}
    \caption{Single perceptron mathematical model \citep{goodfellow2016deep}}
    \label{fig:NN_mo}
\end{figure}

The diagram in \ref{fig:NN_mo} illustrates a conventional mathematical model of a biological neuron, commonly known as a single perceptron. In this model, the concept revolves around making synaptic strengths, denoted by weights $w$, learnable. These weights are pivotal as they determine the extent of influence a neuron has over another, with positive weights indicating excitatory influence and negative weights indicating inhibitory influence. Each neuron in this model receives an input denoted as $(x_{ji})$ and computes the weighted sum of its inputs. This sum is then processed through an activation function to generate the perceptron's output $(o_i)$. Historically, the threshold function was used as an activation function, where the neuron would 'fire' (send a signal) if the calculated sum exceeded a certain threshold. 

In NNs, the frequency of neuron firing transmits information. This firing rate is modeled using an \emph{activation function}. Without such a function, NNs would be akin to linear regression models, lacking the ability to capture complex patterns and non-linear relationships. Several activation functions are used in NNs, each with unique properties. Some of them with their properties are represented in Table \ref{tab:activ}.

\begin{table}[!t]
\centering
\caption{Activation functions and their properties \citep{goodfellow2016deep}}
\label{tab:activ}
\resizebox{.8\columnwidth}{!}{%
\begin{tabular}{|l|l|l|l|} 
\toprule
\multicolumn{1}{|c|}{\textbf{Name}} & \multicolumn{1}{c|}{$\mathbf{f(x)}$} & \multicolumn{1}{c|}{$\mathbf{f^\prime(x)}$} & \multicolumn{1}{c|}{\textbf{Range}}  \\ 
\hline
Identity (Linear)                   & $x$                             & $1$                             & $(-\infty,+\infty)$  \\ 
\hline
\begin{tabular}[c]{@{}l@{}}Binary step\\ (Threshold)\end{tabular} & $\max\{0,1\}=\begin{cases}
1 & \text{if } x\geq 0\\
0 & \text{if } x< 0
\end{cases}$ & $\begin{cases}
0 & \text{if } x\neq 0\\
\text{NaN} & \text{if } x=0
\end{cases}$ & $\{0,1\}$ \\ 
\hline
\begin{tabular}[c]{@{}l@{}}Logistic (sigmoid\\  or soft step)\end{tabular} & $\frac{1}{\mathlarger{1+e^{-x}}}$ & $f( x)( 1-f( x))$ & $( 0,1)$ \\ 
\hline
Tanh & $\frac{\mathlarger{e^{x} -e^{-x}}}{\mathlarger{e^{x} +e^{-x}}}$ & $1-( f( x))^{2}$ & $(-1,1)$ \\ 
\hline
\begin{tabular}[c]{@{}l@{}}Rectified linear\\ unit (ReLU)\end{tabular} & $\max\{x,0\} =\begin{cases}
x & \text{if } x\geq 0\\
0 & \text{if } x< 0
\end{cases}$ & $\begin{cases}
1 & \text{if } x >0\\
0 & \text{if } x< 0\\
\text{NaN} & \text{if } x=0
\end{cases}$ & $[0,+\infty)$ \\ 
\hline
\begin{tabular}[c]{@{}l@{}}Gaussian error \\ linear unit (GELU)\end{tabular} & $x\Phi ( x) =\frac{1}{2} x\left( 1+\text{erf}\left(\frac{1}{\sqrt{2}}\right)\right)$ & $\Phi ( x) +x\Phi ( x)$ & $(-0.17,+\infty)$\\ 
\hline
Softplus & $\ln\left( 1+e^{x}\right)$ & $\frac{1}{\mathlarger{1+e^{-x}}}$ & $(0,+\infty )$ \\ 
\hline
\begin{tabular}[c]{@{}l@{}}Exponential linear\\ unit (ELU)\end{tabular} & $\begin{cases}
\alpha \left( e^{x} -1\right) & \text{if } x\leq 0\\
x & \text{if } x >0
\end{cases}$ & $\begin{cases}
\alpha e^{x} & \text{if } x< 0\\
1 & \text{if } x >0\\
1 & \text{if } x=0
\end{cases}$ & $(-\alpha ,+\infty)$\\ 
\hline
\begin{tabular}[c]{@{}l@{}}Leaky ReLU\end{tabular} & $\begin{cases}
\alpha x & \text{if } x\leq 0\\
x & \text{if } x >0
\end{cases}, 0\leq \alpha < 1$ & $\begin{cases}
\alpha  & \text{if } x\leq 0\\
1 & \text{if } x >0
\end{cases}$ & $(-\infty,+\infty)$\\ 
\hline
\begin{tabular}[c]{@{}l@{}}Sigmoid linear unit \\  (Sigmoid shrinkage)\end{tabular} & $\frac{x}{\mathlarger{1+e^{-x}}}$ & $\frac{\mathlarger{1+e^{-x} +xe^{-x}}}{\mathlarger{\left( 1+e^{-x}\right)^{2}}}$ & $[-0.278,+\infty)$\\
\bottomrule
\end{tabular}}
\end{table}

\subsubsection{Organization of ANNs}

In DL, neurons are commonly arranged into multiple layers. Within a specific layer, neurons establish connections solely with those in the immediately preceding and succeeding layers. This configuration commences with the input layer, designed to receive external data, and concludes with the output layer, responsible for generating the final output. Between these two layers, zero or more hidden layers may exist.

Different patterns of connections exist between layers. One typical pattern is full connectivity, as seen in multi-layer perceptrons (MLP), where each neuron in one layer connects to every neuron in the next layer. Another type is convolutional layers, used in convolutional neural networks (CNNs), where groups of neurons connect to other groups, thereby increasing the depth of the network. These layers form feature maps~\citep{love2023topological}. 

Pooling layers, such as max-pooling or average-pooling, involve clusters of neurons connecting to a single neuron in the subsequent layer, effectively diminishing the number of neurons in that layer. Neurons with exclusively forward connections constitute a directed acyclic graph called feedforward networks. In contrast, networks that permit connections among neurons within the same or preceding layers are recognized as recurrent neural networks (RNNs) \citep{goodfellow2016deep,lim2021understanding,li2022approximation}.

\subsubsection{Training Artificial Neural Networks}

Learning in NNs involves adapting the network to perform tasks using sample observations. This process entails adjusting the network's weights (and, optionally, thresholds) to enhance the accuracy of results. It involves minimizing observed errors, and learning is deemed complete when further analysis of additional observations does not significantly reduce the error rate. It's important to note that the error rate typically doesn't reach zero after completing the learning process. If the error rate remains excessively high post-learning, the network may need restructuring.

The learning process centers around establishing a cost (loss) function regularly assessed during the learning phase. Learning continues as long as the output of this function keeps decreasing. Often described as a statistical measure, the value of the cost function can usually only be approximated. Since the outputs are numerical, a low error indicates minimal difference between the network's output and the correct answer (ground truth). The goal of learning is to minimize these cumulative differences across various observations. Most learning models in NNs apply optimization theory and statistical estimation.

A summary of commonly used cost functions in ANNs is provided in Table~\ref{tab:loss}. In this context, $y$ represents the ground-truth value, $\hat{y}$ is the output of the network's last layer, $\cdot^{(j)}$ denotes $j^{th}$ dimension of a given vector, and $p(\cdot)$ represents a probability estimate.

\begin{table}[!t]
\centering
\caption{List of some of mostly used loss functions \citep{goodfellow2016deep}}
\label{tab:loss}
\begin{tabular}{lll} 
\toprule
\multicolumn{1}{l}{\textbf{Symbol}} & \multicolumn{1}{l}{\textbf{Name}} & \multicolumn{1}{l}{\textbf{Equation}}  \\ 
\hline
$\mathcal{L}_1$ & $\text{L}_1$ Loss & $\left\Vert y-\hat{y}\right\Vert_1$\\
\hline
$\mathcal{L}_2$ & $\text{L}_2$ Loss & $\left\Vert y-\hat{y}\right\Vert_2$\\
\hline
$\mathcal{L}_1\circ p$ & Expectation Loss & $\left\Vert y-p\left(\hat{y}\right)\right\Vert_1$\\
\hline
$\mathcal{L}_2\circ p$ & Regularised Expectation Loss & $\left\Vert y-p\left(\hat{y}\right)\right\Vert_2$\\
\hline
$\mathcal{L}_\infty\circ p$ & Chebyshev Loss & $\left\Vert y-p\left(\hat{y}\right)\right\Vert_\infty$\\
\hline
hinge & Hinge Loss & $\sum_j\max~\left\{0,\frac{1}{2}y^{j}\hat{y}^{j}\right\}$\\
\hline
$\text{hinge}^2$ & Squared Hinge Loss & $\sum_j\max~\left\{0,\frac{1}{2}y^{j}\hat{y}^{j}\right\}^2$\\
\hline
$\text{hinge}^3$ & Cubed Hinge Loss & $\sum_j\max~\left\{0,\frac{1}{2}y^{j}\hat{y}^{j}\right\}^3$\\
\hline
log & Log Loss (cross-entropy)  & $-\sum_{j}y^{(j)}\log p\left(\hat{y}\right)^{(j)}$\\
\hline
$\text{log}^2$ & Squared Log Loss & $-\sum_{j}\left(y^{(j)}\log p\left(\hat{y}\right)^{(j)}\right)^2$\\
\bottomrule
\end{tabular}
\end{table}

Backpropagation (BP) is a crucial algorithm for training the DL models. Efficiently calculates the gradient of the loss function concerning the network weights for each input-output example. This efficiency is significant compared to the less efficient approach of directly computing the gradient for each weight separately. The main benefit of this efficiency is that it makes gradient-based methods, like gradient descent and its variants (e.g., stochastic gradient descent), practical for training the DL models. Using the chain rule, the BP algorithm calculates the gradient of the loss function concerning the weights. It accomplishes this computation systematically, layer by layer \citep{goodfellow2016deep}.

\subsubsection{Trained Artificial Neural Networks}

DNNs, despite their notable successes across various domains, have encountered limited acceptance in safety-critical environments \citep{bunel2018unified,cheng2017maximum}. This hesitation largely stems from the perception of DNNs as "black boxes" with unpredictable behavior. In applications such as autonomous driving and medical image processing, where safety is paramount, it's essential to have methods to verify the dependability of DL models. Traditional methods for evaluating these models typically involve testing with separate, unused datasets, but this might not sufficiently guarantee safety in critical situations.

Recently, there's been a shift towards using formal methods to evaluate trained NNs, explicitly leveraging the piece-wise linear nature of networks using the ReLU activation function. This structural aspect of ReLU-based networks has been pivotal in interpreting FF-DNNs. As \cite{anderson2020strong} notes, if the non-linearities in FF-DNNs are entirely piece-wise linear ReLU functions, the entire network is also piece-wise linear. On the flip side, any continuous piece-wise linear function has the potential to be represented by the NN utilizing ReLU activation functions.

ReLU FF-DNNs can be modeled as high-dimensional piece-wise linear functions to be formulated as MIP. This approach has opened new avenues in researching FF-DNNs, particularly in scenarios where BP algorithms struggle with non-differentiable activation functions like binary step functions. Therefore, this paper presents a comparative study of various MIP formulations of FF-DNNs. The following section will delve into detailed discussions on the MIP formulations for trained and training FF-DNNs.

\section{FF-DNNs as MIP}\label{FF-DNN}

FF-DNNs are structured as networks of neurons connected in an acyclic graph, where neurons between two adjacent layers are fully pairwise connected. Still, neurons within the same layer do not connect. In the case of ReLU-based FF-DNNs, this architecture can be represented as a MIP model. MIP has been widely used in optimizing over-trained ReLU-based FF-DNNs, providing a framework for rigorous analysis of network properties.

For instance, MIP formulation of ReLU-based FF-DNN allows for detailed examination of network behaviors, such as verifying the robustness of output within a constrained input range \citep{lomuscio2017approach,tjeng2018evaluating,bunel2020branch}. Additionally, MIP formulations of ReLU-based FF-DNNs have been employed for various other purposes. These include determining robust perturbation bounds \citep{cheng2017maximum}, count linear regions \citep{bunel2018unified,dutta2017output,serra2018bounding,serra2020empirical}, find adversarial examples \citep{fischetti2018deep,cheng2017maximum}, and identifying critical neurons in the network \citep{elaraby2020identifying}. These applications highlight the versatility of MIP in analyzing and enhancing the performance of FF-DNNs, making it a valuable tool for understanding and improving the reliability and robustness of these NNs.

Table \ref{tab:notation} provides the notations used throughout the section. The output vector $x^L$ of an FF-DNN is calculated by sequentially propagating information from the input layer to each subsequent layer. This propagation involves the calculation of a weighted sum of the inputs along with bias vectors $b^l$ (where $l>0$) and then applying a non-linearity to these sums. 

An activation function, denoted as $\sigma$ (Table \ref{tab:activ}), introduces non-linearity in the network. This function is critical for activating neurons, thereby facilitating the learning process. The equation \eqref{eq:activef} describes this process of non-linearity. It illustrates how the inputs to the neurons of one layer are transformed and subsequently determines the inputs for the following layer. This mechanism is central to how FF-DNNs process and learn from data.

\begin{equation}\label{eq:activef}
    x^l=\sigma\left(w^{l}x^{l-1}+b^{l}\right),\hspace{1cm}\forall{l=1,\dots,L}
\end{equation}
\begin{table}[!t]
\centering
\caption{Notation for FF-DNNs architectures \citep{gambella2020optimization}}
\label{tab:notation}
\begin{tabular}{ll} 
\toprule
\textbf{Notation}                        & \textbf{Description}  \\ 
\hline
$l\in\{0,1,\dots,L\}$                        & Indices of layers  \\
$n^l$                                    & Number of neurons (unit) in each layer $l$\\
$\sigma$                                 & Element-wise activation function \\
$w^l\in\mathbb{R}^{n^{l-1}\times n^{l}}$ & Weight matrix of each layer $l$ \\
$b^l\in\mathbb{R}^{n^l}$                 &  Vector of bias for each layer $l>0$ \\
$(\mathbb{X},\mathbb{Y})$                                  & Dataset for training\\
$x^l$                                    & \begin{tabular}[c]{@{}l@{}}Output vector for each layer $l$; Note that $l=0$ represents the input \\feature vector, $l>0$ represents the derived feature vector.\end{tabular}  \\
\bottomrule
\end{tabular}
\end{table}

Pooling layers in FF-DNNs introduce non-linearity by summarizing the information from the preceding layer. Common pooling layers include Average-Pooling, Global Average-Pooling, Max-Pooling, and Global Max-Pooling. A vital characteristic of these pooling layers is that they do not contain trainable parameters.

During the forward pass in a pooling layer, the layer selects and passes to the next layer a specific value from each pool size. This selection is based on the type of pooling: for instance, Max-Pooling returns the maximum value within each pool, while Average-Pooling returns the average.

In the backward pass, part of the learning process where the network adjusts its weights based on the output error, the pooling layer propagates the error from the subsequent layer back to the original location of the selected value. In Max-Pooling, the error is propagated back to the neuron that contributed the maximum value; in Average-Pooling, the average error is distributed across all neurons in the pool. This backward propagation is crucial for the network to learn effectively, as it allows the network to understand which neurons contributed most significantly to the output and adjust accordingly. 

\begin{figure}[!t]
    \centering
    \includegraphics[width=0.5\textwidth]{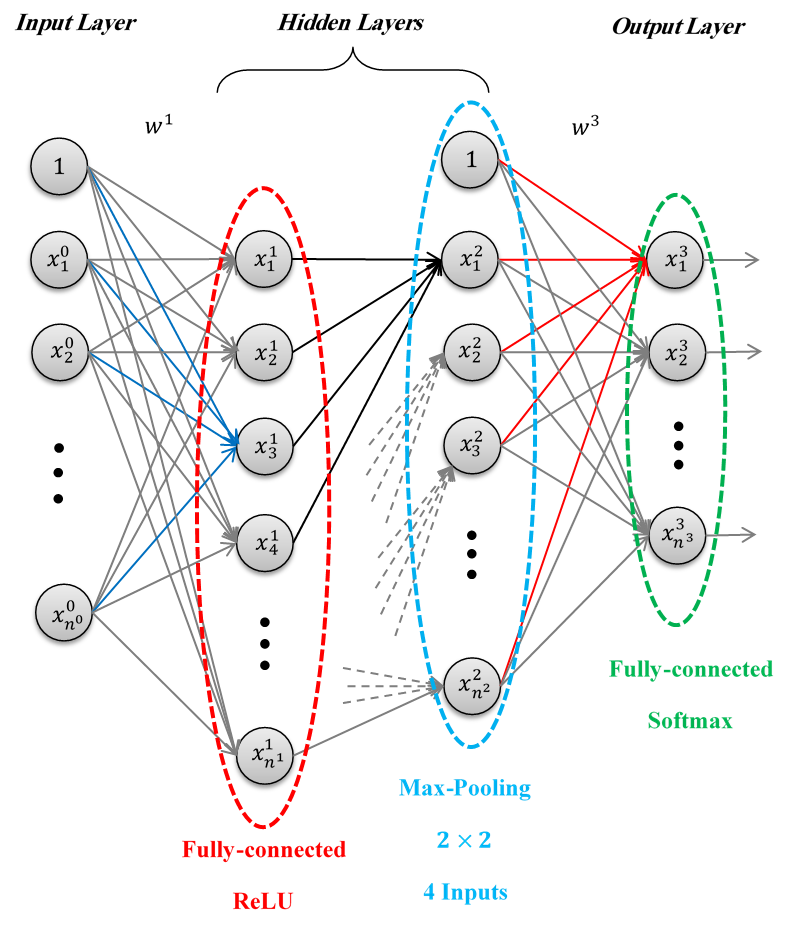}
    \caption{A simple illustration of FF-DNNs, e.g., CNNs}
    \label{fig:DNN_ex}
\end{figure}

Figure \ref{fig:DNN_ex} depicts a three-layer FF-DNNs example $(L=3)$. In this example, the input layer consists of $n^0$ units, the first hidden layer contains $n^1$ units, and the second is a Max-Pooling layer with $n^2$ units. The output layer of the network has $n^3$ units. To train such a network, it is necessary to determine the weight matrices $W^1\in\mathbb{R}^{n^0\times n^1}$, $W^3\in\mathbb{R}^{n^2\times n^3}$, and bias vectors $b^1\in\mathbb{R}^{n^1}$, and $b^3\in\mathbb{R}^{n^3}$. It is important to note that since the second layer is a Max-Pooling layer, it does not have trainable weights. 

In the first layer $(l=1)$, the activation function is ReLU, defined as, $\sigma(x)=\max~\{0,x\}$. Therefore, the output of the units in this layer is given by
\begin{equation}\label{eq:reluout}
    x^1=\max~\left\{0,w^{1}x^{0}+b^{1}\right\}\Longrightarrow x_{i}^{1}=\max~\left\{0,\sum_{j=1}^{n^{0}}w_{ji}^{1}x_{j}^{0}+b_i^1\right\},\hspace{5mm}\forall{i=1,\dots,n^{1}}
\end{equation}
By considering $x_{0}^{l}=1$ in all layers with trainable weights, the bias of layer $l$ is $b^{l}_{i}=w_{0i}^{l}$. This representation allows the weighted sum of the layers to be expressed in matrix form. Hence, Equation \eqref{eq:reluout} can be written as
\begin{equation}\label{eq:reluoutmat}
    x^1=\max~\left\{0,w^{1}x^{0}\right\}\Longrightarrow x_{i}^{1}=\max~\left\{0,\sum_{j=0}^{n^{0}}w_{ji}^{1}x_{j}^{0}\right\},\hspace{5mm}\forall{i=1,\dots,n^{1}}
\end{equation}
In Figure \ref{fig:DNN_ex}, the second layer performs Max-Pooling with a pool size of $2\times2$, considering groups of four inputs. Generally, the output of a Max-Pooling layer with a pool size of $s\times s$ is computed as the maximum value among the inputs in the pool, as described in Equation~\eqref{eq:maxpool}:
\begin{equation}\label{eq:maxpool}
    x_{i}^{l}=\max~\left\{x_{j_1}^{l-1},x_{j_2}^{l-1},\dots,x_{j_{s}}^{l-1}\right\},\hspace{5mm}\forall{i\in\left\{1,\dots,n^l\right\},j\in\left\{1,\dots,n^{l-1}\right\}}
\end{equation}
For instance, $x_{1}^{2}=\max~\left\{x_{1}^{1},x_{2}^{1},x_{3}^{1},x_{4}^{1}\right\}$. The final layer in Figure \ref{fig:DNN_ex} applies the Softmax non-linearity. The Softmax function normalizes the output into a probability distribution over predicted output classes. It can be represented as shown in Equation \eqref{eq:softmaxins}:
\begin{equation}\label{eq:softmaxins}
    x_{k}^{3}=\frac{e^{\sum_{j=0}^{n^2}w_{jk}^{3}x_j^2}}{\sum_{i=1}^{n^3}e^{\sum_{j=0}^{n^2}w_{ji}^{3}x_j^2}},\hspace{5mm}\forall{k=1,\dots,n^3}
\end{equation}
As discussed, ReLU-based FF-DNNs, which employ a piecewise linear model, can be formulated as the MIP. For the remainder of this section, it is assumed that all units in the network are activated via the ReLU non-linearity. Pooling layers, which maintain linearity, are also included in this consideration. This framework allows for the formulation and analysis of the network in terms of MIP, providing a structured approach to understanding and optimizing the network's behavior.

\subsection{Trained DNNs as MIP}\label{ch:2.2}

A FF-DNN is a function $f:\mathbb{R}^{n^0}\longrightarrow\mathbb{R}^{n^L}$ that maps a input $x^0\in\mathbb{R}^{n^0}$ from a high-dimensional space to an output $x^{L}=f\left(x^{0}\right)\in\mathbb{R}^{n^L}$. In a ReLU-based FF-DNN with $L$ layers, the transformation of the input through each layer of the network can be described recursively as Equation \eqref{eq:2.6}.
\begin{equation}\label{eq:2.6}
    x^{l}=\text{ReLU}\left(w^{l}x^{l-1}\right),\hspace{1cm}\forall{l=1,\dots,L}
\end{equation}
alternatively, for each neuron in the network,
\begin{equation}\label{eq:2.7}
    x_{i}^{l}=\text{ReLU}\left(\sum_{j=0}^{n^{l-1}}w_{ji}x_{j}^{l-1}\right),\hspace{5mm}\forall{i=1,\dots,n^{l}\hspace{3mm}l=1,\dots,L}
\end{equation}

Compared to the training of FF-DNNs, which involves optimizing parameters $w$ and $b$, optimizing over FF-DNNs aims to identify extreme cases for a trained model. As a result, model parameters $(w, b)$ are given, and the inputs/outputs of nodes in the network $x^0$ and $x^L$ are the variables instead. As seen in equations \eqref{eq:2.6} and \eqref{eq:2.7}, the piecewise linear function ReLU poses a significant challenge in formulating ReLU-based FF-DNNs as MIP. However, the structural characteristics of ReLU enable it to be represented as a set of linear constraints within MIP models. Four approaches have been proposed in the literature to encode ReLU by linear inequalities. These formulations are crucial because they transform the non-linear ReLU function into a linear form that can be handled by MIP, thus enabling the application of optimization techniques on trained NNs. The remainder of the discussion in this section will delve into these four ReLU formulations. 

\subsubsection{Big-\emph{M} formulation for ReLU} 

The most straightforward approach to model the ReLU function as a set of linear constraints is the big-\emph{M} approach. By introducing a binary variable $z_{i}^{l}$, for $i=1,\dots,n^l$ and $l=1,\dots,L$, that takes $1$ if the neuron is activated, \emph{i.e.}, $\sum_{j=0}^{n^{l-1}}w_{ji}^{l}x_{j}^{l-1}>0$, and $0$ otherwise. For $l=1,\dots,L$, ReLU is formulated as the set of linear inequalities in \eqref{eq:2.8}.
\begin{subequations}\label{eq:2.8}\begin{align}
    x^{l}&\geq0\label{eq:2.8a}\\x^{l}&\geq w^{l}x^{l-1}\label{eq:2.8b}\\x^{l}&\leq w^{l}x^{l-1}+\left(\mathbb{1}-z^{l}\right)\odot {M^{l}}\label{eq:2.8c}\\x^{l}&\leq z^{l}\odot {M^{l}}\label{eq:2.8d}\\x^{0} &\in [L,U]^{n^0}\\x^{l} &\in\mathbb{R}^{n^l}_{+}\\ z &\in \{0,1\}^{n^l}
\end{align}\end{subequations} 
where the symbol $\odot$ in equations \eqref{eq:2.8c} and \eqref{eq:2.8d} denotes the element-wise multiplications, and $\mathbb{1}$ shows the vector of ones.

The efficiency of solving the MIP model with current solvers is significantly influenced by the model's relaxation of Linear Programming (LP) quality. Hence, determining ${M^{l}}$is crucial because it impacts how tightly the LP relaxation estimates the convex hull of the feasible region. One commonly used approach is to establish the bounds ${L^{l}}$ and ${U^{l}}$ for each layer $l$ and propagate the bounds from the input layer to the successive layer through a process known as interval arithmetic. \cite{bunel2018unified} proposed a specific procedure for this propagation, as shown in Equation \eqref{eq:2.9}. This procedure is applied for each layer $l=1,\dots,L$ and each neuron $i=1,\dots,n^l$, to find bounds for $y_{i}^{l}=\sum_{j=0}^{n^{l-1}}w_{ji}^{l}x_{j}^{l-1}$, which in turn determines $x_i^l=\max~\left\{0,y_i^l\right\}$. The propagation of bounds through a ReLU activation essentially applies the ReLU function to the bounds themselves. Subsequently, $M_{i}^{l}=\max\left\{-L_{i}^{l},U_{i}^{l}\right\}$, where
\begin{subequations}\label{eq:2.9}
    \begin{align}
        {L^{l}}_{i} &= \sum_{j=1}^{n^{l-1}}\left({w_{ji}^{l}}^{+}{L_j^{l-1}}^{+}+{w_{ji}^{l}}^{-}{U_j^{l-1}}^{+}\right)+w_{0i}^{l}\\{U^{l}}_{i} &= \sum_{j=1}^{n^{l-1}}\left({w_{ji}^{l}}^{-}{L_j^{l-1}}^{+}+{w_{ji}^{l}}^{+}{U_j^{l-1}}^{+}\right)+w_{0i}^{l}
    \end{align}
\end{subequations}
In these equations, $a^{+}=\max\{a,0\}$ and $a^{-}=\min\{a,0\}$  are used to handle the right-hand side of Equation \eqref{eq:2.9}. 

\cite{cheng2017maximum} proposed different propagating procedure. For $l=1,\dots,L$ and $i=1,\dots,n^l$, they define $y_i^l=\sum_{j=0}^{n^{l-1}}w_{ji}^{l}x_{j}^{l-1}$, and, accordingly, $x_i^l=\max~\left\{0,y_i^l\right\}$. They obtain bounds for the affine transformation of inputs of previous layer $(y_i^l)$ as equations \eqref{eq:2.10}.
\begin{subequations}\label{eq:2.10}
    \begin{align}
        L_{y_i^l} &= \sum_{j=0}^{n^{l-1}}\min~\left\{w_{ji}^{l}L_{x_j^{l-1}},w_{ji}^{l}U_{x_j^{l-1}}\right\}\\ U_{y_i^l} &= \sum_{j=0}^{n^{l-1}}\max~\left\{w_{ji}^{l}L_{x_j^{l-1}},w_{ji}^{l}U_{x_j^{l-1}}\right\}
    \end{align}
\end{subequations}
Obtaining the bounds on the input of each unit, they define,
\begin{equation}
    x_i^l\in\left[L_{i}^{l},U_{i}^{l}\right]=\left[\max\left\{0,L_{y_i^l}\right\},\max\left\{0,U_{y_i^l}\right\}\right]
\end{equation}
If $L_{i}^{l}\geq0$ or $U_{i}^{l}\leq0$, the output of $x_i^l$ is known, and the constraints set \eqref{eq:2.8} is dropped for that neuron. Otherwise, $M_{i}^{l}=U_{i}^{l}$.

\cite{tjeng2018evaluating}pointed out that the interval arithmetic approach, as detailed in Equations \eqref{eq:2.9} and \eqref{eq:2.10}, can result in excessively conservative bounds. To address this, they proposed an alternative method. For each neuron $(x_i^l)$, they construct a sub-tree rooted at $x_i^l$ which corresponds to eliminating unrelated constraints and variables to $x_i^l$ in \eqref{eq:2.8} and maximize (minimize) the value of $x_i^l$ over the LP-relaxation of the remaining set of linear inequalities to obtain upper (lower) bound for $x_i^l$; then, determine ${L^{l}}_{i}$ and ${U^{l}}_{i}$.

\subsubsection*{Convex hull of a single ReLU}

Since the big-\emph{M} method only focuses on determining the value of the ReLU function and not the value of $w^{l}x^{l-1}$, it may not be reliable around the boundary of the domain of this value, while it is generally effective. Anderson et al. (2020) demonstrated that the big-\emph{M} formulation is not tight and introduced a formulation for the convex hull, representing the tightest possible formulation for individual ReLU units \cite{anderson2020strong}. They consider a single ReLU unit with the input $x\in\mathbb{R}^{\eta}$ and the single output $y=\text{ReLU}(wx+b)$. Assuming $x_{i}\in[L_{i},U_{i}]$, for $i=1,\dots,\eta$, they defined,
\begin{equation}
    \Hat{L}_i=\begin{cases}L_i & w_i\geq0\\U_i & w_i<0\end{cases}\hspace{5mm}\text{and}\hspace{5mm}\Hat{U}_i=\begin{cases}U_i & w_i\geq0\\L_i & w_i<0\end{cases}
\end{equation}
Then, they proposed the strongest formulation for a single ReLU unit as a set of linear constraints in \eqref{eq:2.13}.
\begin{subequations}\label{eq:2.13}
    \begin{align}
        y &\leq \sum_{i\in S}w_i\left(x_i-\Hat{L}_{i}(1-z)\right)+\left(b+\sum_{i\notin S}w_{i}\Hat{U}_{i}\right)z && \forall{S\subseteq\textsc{supp}(w)}\\ y &\geq \sum_{i=1}^{\eta}w_{i}x_{i}+b\\ y &\geq 0\\ x &\in [L,U]^{\eta}\\ z &\in \{0,1\}
    \end{align}
\end{subequations}
where
\begin{subequations}
    \begin{align}
    \textsc{supp}(w) &= \left\{i\in\{1,\dots,\eta\}:\arg\min_{z}~\left\{w_{i}x_{i}-w_{i}\left(\Hat{L}_i(1-z)+\Hat{U}_{i}z\right)\right\}=1\right\}\\ &= \left\{i\in\{1,\dots,\eta\}:w_{i}\ne0\right\}
\end{align}
\end{subequations}

In other words, set $S$ denotes the subset of inputs that lead to the activation of neurons. They argue that when applying formulation \eqref{eq:2.13} to an entire NN, the specific properties that make the formulation effective for individual components may not fully apply to the combined network model. Nonetheless, when dealing with formulations of comparable size, the combination of stronger formulations typically results in more computationally efficient complete formulations.

Comparing \eqref{eq:2.8} and \eqref{eq:2.13} reveals that instead of featuring two inequalities per neuron in each layer, namely \eqref{eq:2.8c} and \eqref{eq:2.8d}, the convex hull formulations necessitate either an exponential number of constraints concerning node inputs or the introduction of numerous additional/auxiliary variables.

\subsubsection{Extended formulation for ReLU} 

Given $x^{l}=\max\left\{0,w^{l}x^{l-1}\right\}$, an extended formulation for ReLU may be generated by introducing non-negative auxiliary continuous variables. In this way, the ReLU formulation is lifted into a space containing the output of a complementary set of neurons, each active when the corresponding neuron is not. By defining $\Bar{x}^{l}=\max\left\{0,-w^{l}x^{l}\right\}$ in each layer $l=1,\dots,L$, such that,
\begin{equation}\label{eq:2.14}
    w^{l}x^{l-1}=x^{l}-\Bar{x}^{l},\hspace{1cm}x^{l}\geq0,\Bar{x}^{l}\geq0
\end{equation}
to decouple the positive and negative parts of the ReLU input. To obtain the unique solution to constraints \eqref{eq:2.14}, at least one of the two terms $x^{l}$ and $\Bar{x}^{l}$ must be zero. Therefore, a binary variable $z^{l}_{i}$, per unit in each layer, is introduced that takes $1$ if $\Bar{x}_{i}^{l}\leq0$ and $0$ if $x^{l}_{i}\leq0$. Supposing that finite non-negative values ${M^l}^{+}$ and ${M^l}^{-}$ could be computed, such that $-{M^l}^{-}\leq w^{l}x^{l-1}\leq {M^l}^{+}$, these indicator constraints are internally converted into appropriate linear inequalities,
\begin{equation}\label{eq:2.15}
    x^{l}\leq {M^l}^{+}z^{l},\quad \Bar{x}^{l}\leq {M^l}^{-}\left(1-z^{l}\right)
\end{equation}
For $l=1,\dots,L$, the constraint sets \eqref{eq:2.16} denote the extended formulation of ReLU.
\begin{subequations}\label{eq:2.16}
    \begin{align}
        w^{l}x^{l-1} &= x^{l}-\Bar{x}^{l}\\x^{l} &\leq {M^l}^{+}z^{l}\\ \Bar{x}^{l} &\leq {M^l}^{-}\left(1-z^{l}\right)\\ x^{0} &\in \left[L^0,U^0\right]^{n^0}\\ x^{l} &\in \left[L^l,U^l\right]^{n^l} \\ \Bar{x}^{l} &\in \left[\Bar{L}^l,\Bar{U}^l\right]^{n^l} \\ z^{l} &\in \{0,1\}^{n^l}
    \end{align}
\end{subequations}

Same as the big-\emph{M} method, determining the bounds on the variables, and, consequently, ${M^l}^{+}$ and ${M^l}^{-}$ have significant impact on LP-relaxation of this formulation. \cite{fischetti2018deep} and \cite{kumar2019equivalent} applied a similar approach as \cite{tjeng2018evaluating}, which is discussed in the previous section to determine the bound. They remove unrelated constraints and variables to $x_i^l$, then solve the obtained MIP (without LP-relaxation) twice: (1) maximizing the value of $x_i^l$, and (2) maximizing the value of $\Bar{x}_i^l$, to determine the bounds.

As another approach, \cite{serra2018bounding} proposed a propagation-based paradigm to obtain the bounds of \eqref{eq:2.16}. Assuming the lower and upper bound of $x^0$ (i.e., ${M^0}^{+}$ and ${M^0}^{-}$) are available, they compute for $l=1,\dots,L$ and $i=1,\dots,n_{l}$
\begin{subequations}\label{eq:2.17}
    \begin{align}
        {M^l}^{+}_{i} &= \max~\left\{0,\sum_{j=1}^{n^{l-1}}\max~\left\{0,w_{ji}^{l}{M^{l-1}}^{+}_{j}\right\}+w_{0i}^{l}\right\}\label{eq:2.21a}\\{M^l}^{-}_{i} &= \max~\left\{0,\sum_{j=1}^{n^{l-1}}\max~\left\{0,-w_{ji}^{l}{M^{l-1}}^{+}_{j}\right\}-w_{0i}^{l}\right\}\label{eq:2.21b}
    \end{align}
\end{subequations}

\subsubsection*{Valid inequalities for the extended formulation of ReLU}

As we discussed in big-\emph{M} method, while determining the bounds for $x_i^l$, if $L_{x_i^l}\geq0$ or $U_{x_i^l}\leq0$, the output of $x_i^l$ is known to be $x_i^l=\sum_{j=0}^{n^{l-1}}w_{ji}x_j^{l-1}$ or $x_i^l=0$, respectively. Hence, if ${M^l}^{+}_{i}<0$, for some $i$ in layer $l$, the unit is always inactive, denoted as stably inactive, and we can remove such units from the formulation. Similarly, if ${M^l}^{-}_{i}<0$, for some $i$ in layer $l$, the unit is always active, denoted as stably active, and we can replace constraints \eqref{eq:2.16} with $x^l=w^{l}x^{l-1}$. In certain large networks, many units are stable \citep{serra2020empirical}. The remaining units, where activity depends on the input, are denoted unstable.

Motivated by this fact in the extended formulation, \cite{serra2020empirical} proposed valid inequalities for the extended formulation. For a unit $i$ in layer $l$ to be active when $w_{0i}^l\leq0$, \emph{i.e.}, negative bias, there must be a positive contribution (active unit). Thus, some unit $j$ in layer $l-1$ must be existed, such that, $w_{ji}^{l}>0$. Therefore, for each $l=2,\dots,L$ and $i=1,\dots,n^{l}$, for which, $w_{0i}^l\leq0$, they proposed the valid inequality in \eqref{eq:ve1}.
\begin{equation}\label{eq:ve1}
    z_{i}^{l}\leq\sum_{j\in\left\{1,\dots,n^{l-1}\right\}:w_{ji}^{l}>0} z_{j}^{l-1}
\end{equation}

Similarly, unit $i$ is only inactive when $w_{0i}^l\geq0$, and, for some active unit $j$ in layer $l-1$, $w_{ji}^{l}<0$. Therefore, for each $l=2,\dots,L$ and $i=1,\dots,n^{l}$, for which, $w_{0i}^l\geq0$, , they proposed the valid inequality in \eqref{eq:ve2}.
\begin{equation}\label{eq:ve2}
    \left(1-z_{i}^{l}\right)\leq\sum_{j\in\left\{1,\dots,n^{l-1}\right\}:w_{ji}^{l}<0} z_{j}^{l-1}
\end{equation}

\subsubsection{Disjunctive Programming for ReLU}

Each ReLU unit can be alternatively modeled as a disjunctive program. For $l=1,\dots,L$ and $i=1,\dots,n^{l}$, ReLU is equivalent to
\begin{equation}\label{eq:2.20}
    \left[\begin{matrix}x_{i}^{l}=0\\ \sum_{j=1}^{n^{l-1}}w_{ji}^{l}x_{j}^{l-1}+b_{i}^{l}\leq0\end{matrix}\right]\bigvee\left[\begin{matrix}x_{i}^{l}=\sum_{j=1}^{n^{l-1}}w_{ji}^{l}x_{j}^{l-1}+b_{i}^{l} \\ \sum_{j=1}^{n^{l-1}}w_{ji}^{l}x_{j}^{l-1}+b_{i}^{l}\geq0\end{matrix}\right]
\end{equation}

The extended formulation for the disjunctive program presents the auxiliary variables for each disjunction. Defining $y_{i}^{l}:=\sum_{j=1}^{n^{l-1}}w_{ji}^{l}x_{j}^{l-1}$, ${y_{i}^{l}}^{a}\in\mathbb{R}$ and ${y_{i}^{l}}^{b}\in\mathbb{R}$, for $l=1,\dots,L$ and $i=1,\dots,n^{l}$, \eqref{eq:2.20} can be equivalently formulated as \eqref{eq:2.21} \citep{anderson2020strong,tsay2021partition}:
\begin{subequations}\label{eq:2.21}
    \begin{align}
        \sum_{j=1}^{n^{l-1}}w_{ji}^{l}x_{j}^{l-1} &= {y_{i}^{l}}^{a}+{y_{i}^{l}}^{b}\\{y_{i}^{l}}^{a}+z_{i}^{l}b_{i}^{l} &\leq 0 \\ {y_{i}^{l}}^{b}+\left(1-z_{i}^{l}\right)b_{i}^{l} &\geq 0 \\ {y_{i}^{l}}^{b}+\left(1-z_{i}^{l}\right)b_{i}^{l} &= x_{i}^{l}\\z_{i}^{l}{L_{i}^{l}}^{a}\leq{y_{i}^{l}}^{a} &\leq z_{i}^{l}{U_{i}^{l}}^{a} \\ \left(1-z_{i}^{l}\right){L_{i}^{l}}^{b}\leq{y_{i}^{l}}^{b} &\leq \left(1-z_{i}^{l}\right){U_{i}^{l}}^{b}\\z_{i}^{l} &\in \{0,1\}
    \end{align}
\end{subequations}

Like the previous formulations, determining bounds in \eqref{eq:2.21} is significant for having a tight and strong LP relaxation. \cite{tsay2021partition} proposed partitioning inputs of each neuron and forming the convex hull over them. In particular, the input nodes in layer $l-1$ are partitioned into subsets $\mathbb{S}_{1}^{l-1}\cup\mathbb{S}_{2}^{l-1}\cup\dots\cup\mathbb{S}_{K}^{l-1}=\{1,\dots,n^{l-1}\}$; $\mathbb{S}_{k}^{l-1}\cap\mathbb{S}_{k^\prime}^{l-1}=\emptyset$ and $k\ne k^\prime$. For every partition, An auxiliary variable is presented, \emph{i.e.}, $y_{i_{k}}^{l}=\sum_{j\in\mathbb{S}_{k}^{l-1}}w_{ji}^{l}x_{j}^{l-1}$. Replacing $y_{i}^{l}:=\sum_{j=1}^{n^{l-1}}w_{ji}^{l}x_{j}^{l-1}$ with $\sum_{k=1}^{K}y_{i_{k}}^{l}$, the disjunctive formulation in \eqref{eq:2.20} becomes:
\begin{equation}\label{eq:2.22}
    \left[\begin{matrix}x_{i}^{l}=0\\ \sum_{k=1}^{K}y_{i_{k}}^{l}+b_{i}^{l}\leq0\end{matrix}\right]\bigvee\left[\begin{matrix}x_{i}^{l}=\sum_{k=1}^{K}y_{i_{k}}^{l}+b_{i}^{l} \\ \sum_{k=1}^{K}y_{i_{k}}^{l}+b_{i}^{l}\geq0\end{matrix}\right]
\end{equation}

Then, for $l=1,\dots,L$ and $i=1,\dots,n^{l}$, the extended formulation then introduces the auxiliary variables ${y_{i_{k}}^{l}}^{a}$ and ${y_{i_{k}}^{l}}^{b}$ for each $y_{i_{k}}^{l}$:
\begin{subequations}
    \begin{align}
        \sum_{j\in\mathbb{S}_{k}^{l-1}}w_{ji}^{l}x_{j}^{l-1} &= {y_{i_{k}}^{l}}^{a}+{y_{i_{k}}^{l}}^{b} && \forall{k=1,\dots,K}\\ \sum_{k=1}^{K}{y_{i_{k}}^{l}}^{a}+z_{i}^{l}b_{i}^{l} &\leq 0\\ \sum_{k=1}^{K}{y_{i_{k}}^{l}}^{b}+\left(1-z_{i}^{l}\right)b_{i}^{l} &\geq 0 \\ \sum_{k=1}^{K}{y_{i_{k}}^{l}}^{b}+\left(1-z_{i}^{l}\right)b_{i}^{l} &= x_{i}^{l} \\ z_{i}^{l}{L_{i_{k}}^{l}}^{a}\leq{y_{i_{k}}^{l}}^{a} &\leq z_{i}^{l}{U_{i_{k}}^{l}}^{a} && \forall{k=1,\dots,K}\\ \left(1-z_{i}^{l}\right){L_{i_{k}}^{l}}^{b}\leq{y_{i_{k}}^{l}}^{b} &\leq \left(1-z_{i}^{l}\right){U_{i_{k}}^{l}}^{b} && \forall{k=1,\dots,K}\\ z_{i}^{l}\in\{0,1\}
    \end{align}
\end{subequations}
To obtain bounds for ${y_{i_{k}}^{l}}^{a}$ and ${y_{i_{k}}^{l}}^{b}$, for each neuron in layer $l$, they solved $4K$ LP-relaxed formulation of MIP: (1) maximizing $\sum_{j\in\mathbb{S}_{k}^{l-1}}w_{ji}^{l}x_{j}^{l-1}$, (2) minimizing $\sum_{j\in\mathbb{S}_{k}^{l-1}}w_{ji}^{l}x_{j}^{l-1}$. They applied an equal-size and equal-range partitioning strategy in which equal-size partitioning was more promising.

\subsubsection{Pooling layer formulations}

Beyond ReLU, pooling layers are prevalent in forming ReLU-FF-DNNs. The typical pooling layers are Max-Pooling and Average-Pooling, whose formulations are discussed in this section.

\subsubsection*{Max-Pooling}

Given a Max-Pooling function $$x_{i}^{l}=\max~\{x_1^{l-1},\dots,x_k^{l-1}\}$$ introducing binary variables $\delta_i$, $i=1,\dots,k$, that takes $1$ if $x_{j}^{l}=x_i^{l-1}$, and $0$ otherwise, Max-Pooling function is equivalently formulated as,
\begin{subequations}
    \begin{align}
        x_{i}^{l} &\geq x_j^{l-1} && &\forall{i\in\{1,\dots,k\}}\\ x_{i}^{l} &\leq x_j^{l-1}+(U_{max}-L_i)(1-\delta_i) && &\forall{i\in\{1,\dots,k\}}\\\sum_{i=1}^{k}\delta_i &= 1 
    \end{align}
\end{subequations}
where $U_{max}=\max~\left\{U_{x_1^{l-1}},\dots,U_{x_k^{l-1}}\right\}$ \citep{cheng2017maximum,bunel2018unified,fischetti2018deep,anderson2020strong}.

\subsubsection*{Average-Pooling}

Average Pooling denoted in \eqref{eq:2.25} is linear by definition and does not require extra efforts \citep{cheng2017maximum,fischetti2018deep}.
\begin{equation}\label{eq:2.25}
    x_{i}^{l}=AvgPool\left(x_{1}^{l-1},x_{2}^{l-1},\dots,x_{k}^{l-1}\right)=\frac{1}{k}\sum_{j=1}^{k}x_{j}^{l-1}
\end{equation}

\subsection{Training DNNs by solving a MIP}

Recent studies indicate utilizing MIP for modeling FF-DNNs and training them with discrete optimization solvers could be effective mainly when working with minimal data \citep{thorbjarnarson2021training,icarte2019training}. The effectiveness of MIP solvers with large NNs and extensive datasets is still debatable. Nonetheless, they show potential in training small networks with small data batches. Training with MIP solvers might not reach the performance levels of gradient-based methods, but it offers a notable advantage by eliminating the need for extensive hyperparameter tuning. In MIP-based training, parameters such as learning rates, momentum, decay, number of epochs, and batch sizes become less critical. MIP models of FF-DNNs, with well-defined objective functions, could theoretically find optimal solutions, a clarity often lacking in gradient-based methods despite thorough hyperparameter tuning.

Training a DNN involves finding the optimal weights $w^l$ and biases $b^l$ that enable the model to fit the training data accurately. This process is guided by a specific measure of training loss $(\mathcal{L})$, which varies depending on the intended purpose or objective of the model. The corresponding optimization problem for training a DNN is as follows. In this formulation, the set $\mathcal{C}$ depends on the type of the activation function.
\begin{mini!}|s|
{x,w}
{\mathcal{L}(y,x^L)}
{\label{eq:2.26}}{}
\addConstraint{x^l}{= \label{eq:2.26b}}{\sigma\left(w^{l}x^{l-1}\right)\quad}{\forall{l=1,\dots,L}}
\addConstraint{w^l}{\in \label{eq:2.26c}}{\mathbb{R}^{n^{l-1}\times n^{l}}\hspace{1.05cm}}{\forall{l=1,\dots,L}}
\addConstraint{x^l}{\in\label{eq:2.26d}}{\mathcal{C}^{n^{l}}\hspace{2.3cm}}{\forall{l=1,\dots,L}}
\addConstraint{x^{0}}{\in}{\mathbb{X}}
\addConstraint{y}{\in}{\mathbb{Y}}
\end{mini!}

Recalling Table \ref{tab:activ}, some activation functions are discrete, making standard BP fail on training FF-DNNs model with such an activation function. Also, recently, some FF-DNNs with discrete structures have emerged for particular purposes that require specific training paradigms. In the remaining, we study the proposed MIPs for these FF-DNNs.

\subsubsection{Binary FF-DNNs}

Binary FF-DNNs refer to classical DNN with binary activation functions \citep{goodfellow2016deep}:
\begin{equation}
    \sigma(x)=\begin{cases}1 & x\geq0 \\ 0 & x<0\end{cases}
\end{equation}

\cite{bah2020integer} encoded the problem \eqref{eq:2.26} as MIP for training purposes. They assume a binary activation function activates all neurons. First, they obtain a mixed-integer non-linear program (MINLP) formulation by formulating \eqref{eq:2.26b} as a set of linear inequalities. Since $x^{l}\in\{0,1\}^{n^{l}}$, 

\begin{equation}\label{eq:2.28}
    w^{l}x^{l-1}<M^{l}x^{l},\hspace{1cm}w^{l}x^{l-1}\geq -M^{l}\left(\mathbb{1}-x^{l}\right),\hspace{1cm} \forall{l=1,\dots,L}
\end{equation}
In this formulation, they assume $w^{l}\in[-1,+1]^{n^{l-1}\times n^{l}}$ and $x^{0}\in[-r,r]^{n^{0}}$; then, $M^{1}=nr+1$ $(x^{0}\in\mathbb{X}\subseteq\mathbb{R}^{n})$ and $M^{l}=n^{l-1}+1$. The formulation \eqref{eq:2.28} is MINLP due to the multiplication of a real-valued variable $w^{l}$ to a binary variable $x^{l-1}$, which forms a bi-linear function. To re-formulate \eqref{eq:2.28} as a MIP, they define
\begin{equation}\label{eq:2.29}
    u^{l}=w^{l}x^{l-1}\Longrightarrow u_{i}^{l}=\sum_{j=0}^{n^{l-1}}w_{ji}^{l}x_{j}^{l},\hspace{1cm}\forall{l=2,\dots,L,\hspace{2mm}i=1,\dots,n^{l}}
\end{equation}
For each term of the summation in \eqref{eq:2.29}, define
\begin{equation}\label{eq:2.30}
    u_{ji}^{l}=w_{ji}^{l}x_{j}^{l-1},\hspace{1cm}\forall{l=1,\dots,L,\hspace{2mm}i=1,\dots,n^{l},\hspace{2mm}j=1,\dots,n^{l-1}}
\end{equation}
To ensure \eqref{eq:2.30} holds, for $l=2,\dots,L$, $i=1,\dots,n^{l}$, and $j=1,\dots,n^{l-1}$, they define the following set of inequalities,
\begin{subequations}\label{eq:2.31}
    \begin{align}
        u_{ji}^{l} &\leq x_{j}^{l-1}\\
        u_{ji}^{l} &\geq -x_{j}^{l-1}\\
        u_{ji}^{l} &\leq w_{ji}^{l}+\left(1-x_{j}^{l-1}\right)\\
        u_{ji}^{l} &\geq w_{ji}^{l}-\left(1-x_{j}^{l-1}\right)
    \end{align}
\end{subequations}
Therefore, for $l=2,\dots,L$ and $i = 1,\dots,n^{l}$, \eqref{eq:2.28} is equivalent to,
\begin{equation}
    \sum_{j=0}^{n^{l-1}}u_{ji}^{l}<M^{l}x_{i}^{l},\hspace{1cm} \sum_{j=0}^{n^{l-1}}u_{ji}^{l}\geq-M^{l}\left(1-x_{i}^{l}\right)
\end{equation}
Finally, for binary FF-DNNs, the constraints set of \eqref{eq:2.26} is equivalently defined as the following,
\begin{subequations}\label{eq:2.33}
    \begin{align}
         \sum_{j=0}^{n^0}w_{ji}^{1}x_{j}^{0} &< M^{1}x_{i}^{1} && \forall i=1,\dots,n^{1}\\
        \sum_{j=0}^{n^0}w_{ji}^{1}x_{j}^{0} &\geq -M^{1}\left(1-x_{i}^{1}\right) && \forall i=1,\dots,n^{1}\\
        \sum_{j=0}^{n^{l-1}}u_{ji}^{l} &< M^{l}x_{i}^{l} && \forall l =2,\dots,L, \hspace{2mm}i = 1,\dots,n^{l} \\
        \sum_{j=0}^{n^{l-1}}u_{ji}^{l} &\geq -M^{l}\left(1-x_{i}^{l}\right) && \forall l =2,\dots,L, \hspace{2mm}i = 1,\dots,n^{l}\\
        u_{ji}^{l} &\leq x_{j}^{l-1} && \forall l=2,\dots,L,\hspace{2mm}i=1,\dots,n^{l},\hspace{2mm}j=0,\dots,n^{l-1}\\
        u_{ji}^{l} &\geq -x_{j}^{l-1} && \forall l=2,\dots,L,\hspace{2mm}i=1,\dots,n^{l},\hspace{2mm}j=0,\dots,n^{l-1}\\
        u_{ji}^{l} &\leq w_{ji}^{l}+\left(1-x_{j}^{l-1}\right)&& \forall l=2,\dots,L,\hspace{2mm}i=1,\dots,n^{l},\hspace{2mm}j=0,\dots,n^{l-1}\\
        u_{ji}^{l} &\geq w_{ji}^{l}-\left(1-x_{j}^{l-1}\right) && \forall l=2,\dots,L,\hspace{2mm}i=1,\dots,n^{l},\hspace{2mm}j=0,\dots,n^{l-1}\\
        x_{i}^{l} &\in \{0,1\} && \forall l=1,\dots,L,\hspace{2mm}i=1,\dots,n^{l}\\
        w_{ji}^{l} &\in [-1,+1] && \forall l=1,\dots,L,\hspace{2mm}i=1,\dots,n^{l},\hspace{2mm}j=0,\dots,n^{l-1}\\
        u_{ji}^{l} &\in [-1,+1] && \forall l=2,\dots,L,\hspace{2mm}i=1,\dots,n^{l},\hspace{2mm}j=0,\dots,n^{l-1}
    \end{align}
\end{subequations}
If the learning task is regression, we can set $x^{L}=w^{L}x^{l-1}$. The training MIP model is based on linear inequalities, as described in \eqref{eq:2.33}. It is applied to each sample in the training dataset. Depending on the specific learning task and the chosen loss function, this formulation can be a standard MIP, a mixed-integer quadratic program (MIQP), or a mixed-integer nonlinear program (MINLP).

\subsubsection{Binarized Neural Networks}

Hubara et al. (2016) demonstrated in a recent study that Binarized Neural Networks (BNNs), where weights and activations are confined to $\{-1, +1\}$, exhibit comparable test performance to standard FF-DNNs in two widely recognized image recognition datasets \cite{hubara2016binarized}. Based on BNNs, the constraints \eqref{eq:2.26c} and \eqref{eq:2.26d} are $w^l\in\{-1,0,1\}^{n^{l-1}\times n^{l}}$ and $x^l\in\{-1,1\}^{n^l}$, respectively. The activation function in \eqref{eq:2.26b} is 
\begin{equation}
    \sigma(x)=\begin{cases}+1 & x\geq0 \\ -1 & x<0\end{cases}
\end{equation}

\cite{icarte2019training} and  \cite{thorbjarnarson2021training} formulated the problem \eqref{eq:2.26} as MIP. In contrast to BNNs, they aimed to enhance the utilization of data and train networks with non-binary integer-valued parameters, \emph{i.e.},$w^l\in\{-P,0,P\}^{n^{l-1}\times n^{l}}$ and $x^l\in\{-P,P\}^{n^l}$ where $P\in\mathbb{Z}_+$.

To formulate inequality \eqref{eq:2.26b}, they defined
\begin{equation}\label{eq:2.35}
    u^{l}=w^{l}x^{l-1}\Longrightarrow u_{i}^{l}=\sum_{j=0}^{n^{l-1}}w_{ji}^{l}x_{j}^{l-1},\hspace{5mm}\forall{l=1,\dots,L,\hspace{5mm}i=1,\dots,n^l}
\end{equation}
For each term of the summation in \eqref{eq:2.35}, define
\begin{equation}\label{eq:2.36}
    u_{ji}^{l}=w_{ji}^{l}x_{j}^{l-1},\hspace{1cm}\forall{l=1,\dots,L,\hspace{3mm}i=1,\dots,n^{l},\hspace{3mm}j=1,\dots,n^{l-1}}
\end{equation}
Substituting \eqref{eq:2.36} in \eqref{eq:2.35}, 
\begin{equation}
    u_{i}^{l}=\sum_{j=1}^{n^{l-1}}u_{ji}^{l},\hspace{1cm}\forall{l=1,\dots,L,\hspace{3mm}i=1,\dots,n^{l}}
\end{equation}
A binary variable $z_{i}^{l}$ is defined, for $l=1,\dots,L-1$ and $i=1,\dots,n^l$, that takes $1$ if $i^{th}$ neuron on $l^{th}$ layer is active, \emph{i.e.}, $\sum_{j=1}^{n^{l-1}}u_{ji}^{l}\geq0$, and $0$ if it is not active. Hence, applying big-\emph{M} method,
\begin{equation}
    -M\left(1-z_{i}^{l}\right)\leq\sum_{j=1}^{n^{l-1}}u_{ji}^{l}\leq M z_{i}^{l},\hspace{1cm}\forall{l=1,\dots,L-1,\hspace{3mm}i=1,\dots,n^{l}}
\end{equation}
By this definition, the activation function is formulated as
\begin{equation}\label{eq:2.39}
    x_{i}^{l}=\left(2z_{i}^{l}-1\right)P,\hspace{1cm}\forall{l=1,\dots,L-1,\hspace{3mm}i=1,\dots,n^{l}}
\end{equation}
Substituting relation \eqref{eq:2.39} into \eqref{eq:2.36},
\begin{equation}\label{eq:2.40}
    u_{ji}^{l}=w_{ji}^{l}\left(2z_{j}^{l-1}-1\right)P,\hspace{1cm}\forall{l=1,\dots,L-1,\hspace{3mm}i=1,\dots,n^{l},\hspace{3mm}j=1,\dots,n^{l-1}}
\end{equation}
For $l=2,\dots,L-1$, $i=1,\dots,n^{l}$, and $j=0,1,\dots,n^{l-1}$, relation \eqref{eq:2.40} is re-formulated as the following linear inequalities,
\begin{subequations}
    \begin{align}
        u_{ji}^{l}-w_{ji}^{l}+2P z_{j}^{l-1} &\leq 2P \\
        u_{ji}^{l}+w_{ji}^{l}-2P z_{j}^{l-1} &\leq 0 \\
        u_{ji}^{l}-w_{ji}^{l}-2P z_{j}^{l-1} &\geq -2P \\
        u_{ji}^{l}+w_{ji}^{l}+2P z_{j}^{l-1} &\geq 0
    \end{align}
\end{subequations}
Since input is the training data $x^{0}\in\mathbb{X}$, for $l=1$, $u_{ji}^1=w_{ji}^{1}x_{j}^{0}$; and, for the output layer,
\begin{equation}
    x_{i}^{L}=\frac{2}{P\left(n^{L-1}+1\right)}\sum_{j=0}^{n^{l-1}}u_{ji}^{l},\hspace{1cm}\forall{i=1,\dots,n^{L}}
\end{equation}
Finally, the constraints set of the training model \eqref{eq:2.26} is as the following,
\begin{subequations}\label{eq:2.43}
    \begin{align}
    -M\left(1-z_{i}^{l}\right)\leq\sum_{j=1}^{n^{l-1}}u_{ji}^{l} &\leq M z_{i}^{l} && \forall{l\in L\setminus\{L\},\hspace{0.56cm}i\in n^{l}}\\
    u_{ji}^1 &= w_{ji}^{1}x_{j}^{0} && \forall{j\in n^{0},\hspace{1.55cm}i\in n^{1}}\\
    u_{ji}^{l}-w_{ji}^{l}+2P z_{j}^{l-1} &\leq 2P && \forall{l\in L\setminus\{L,1\},\hspace{2mm}i\in n^{l},\hspace{2mm}j\in n^{l-1}} \\
    u_{ji}^{l}+w_{ji}^{l}-2P z_{j}^{l-1} &\leq 0 && \forall{l\in L\setminus\{L,1\},\hspace{2mm}i\in n^{l},\hspace{2mm}j\in n^{l-1}} \\
    u_{ji}^{l}-w_{ji}^{l}-2P z_{j}^{l-1} &\geq -2P && \forall{l\in L\setminus\{L,1\},\hspace{2mm}i\in n^{l},\hspace{2mm}j\in n^{l-1}} \\
    u_{ji}^{l}+w_{ji}^{l}+2P z_{j}^{l-1} &\geq 0 && \forall{l\in L\setminus\{L,1\},\hspace{2mm}i\in n^{l},\hspace{2mm}j\in n^{l-1}} \\
    \frac{2}{P\left(n^{L-1}+1\right)}\sum_{j=0}^{n^{l-1}}u_{ji}^{l} &= x_{i}^{L} && \forall{i\in n^{L}} \\
    u_{ji}^{l} &\in \mathbb{R} && \forall{l\in L,\hspace{2mm}i\in n^{l},\hspace{2mm}j\in n^{l-1}} \\
    z_{i}^{l} &\in \{0,1\} && \forall{l\in L\setminus\{L\},\hspace{2mm}i\in n^{l}} \\
    w_{ji}^{l} &\in \{-P,0,P\} && \forall{l\in L,\hspace{2mm}i\in n^{l},\hspace{2mm}j\in n^{l-1}}
    \end{align}
\end{subequations}
When $P=1$, the formulation in \eqref{eq:2.43} is equivalent to the formulation proposed by \cite{icarte2019training}. Solving BDNNs to optimality, however, remains an unresolved challenge in the field \citep{bah2020integer}

\section{Computational experiments}\label{experiments}

In conclusion, we conduct an initial computational study to explore various formulation approaches for networks trained with ReLU. We focus on trained FF-DNNs because of various formulations for their MIP formulation. The reader must note that the formulations for trained FF-DNNs and training them differ. In other words, given training dataset $(\mathbb{X},\mathbb{Y})$, the aim of training FF-DNNs is to find weights $(w)$ and biases $(b)$ of the function $f_{w,b}:\mathbb{R}^{n\times m}\longrightarrow\mathbb{R}^{k}$, such that minimize a loss function $\mathcal{L}\left(\mathbb{Y},f_{w,b}\left(\mathbb{X}\right)\right)$. In this case, $w$'s and $b$'s along with the output of hidden neurons $x$'s are variable. Therefore, ReLU formulations in Section \ref{ch:2.2} are not applicable for training FF-DNNs since we have a non-linearity $wx$ in the formulation. Hence, One needs to obtain a proper formulation for ReLU units for training purposes.

\subsection{Experimental setup}
We address a conventional classification problem, specifically hand-written digit recognition of a $28\times28$ image. Each of the $784$ pixels in the image is normalized to represent a gray level on a $[0,1]$ scale, where $1$ signifies white and $0$ denotes black. The MNIST dataset \citep{lecun1998gradient} serves as the training dataset for our network, aiming to classify digits from "0" to "9".

For the computational study, we consider two network designs for the remarked classification task. The details of each design are illustrated in Table \ref{tab:set}. The depth of both networks is the same, but Network 2 is more expansive than Network 1. Both networks are trained by TensorFlow \citep{abadi2016tensorflow} up to $500$ epochs with early-stopping technique, and learning rate $1\text{e}-4$ with Adam optimizer.

\begin{table}[!b]
\centering
\caption{Experimental set up of architectures for FF-DNNs}
\label{tab:set}
\resizebox{\columnwidth}{!}{%
\begin{tabular}{|c|c|cccccccc|ccc|} 
\hline
\multirow{2}{*}{\textbf{Architecture}} & \multirow{2}{*}{\textbf{Layers}} & \multicolumn{8}{c|}{\textbf{Number of Neurons}} & \multicolumn{3}{c|}{\textbf{Training Results (accuracy)}}  \\ 
\cline{3-13}
&   & \textbf{Input} &\textbf{H\textsubscript{1}} & \textbf{H\textsubscript{2}} & \textbf{H\textsubscript{3}} & \textbf{H\textsubscript{4}} & \textbf{H\textsubscript{5}} & \textbf{H\textsubscript{6}} & \textbf{Output} & \textbf{Training} & \textbf{Validation} & \textbf{Test}    \\ 
\hline
Network 1                              & $7$                              & $784$          & $20$              & $20$              & $10$              & $10$              & $10$              & $10$              & $10$            & $96.42\%$         & $91.72\%$           & $94.27\%$        \\
Network 2                              & $7$                              & $748$          & $256$             & $128$             & $64$              & $32$              & $16$              & $10$              & $10$            & $99.14\%$         & $96.05\%$           & $97.1\%$         \\
\hline
\end{tabular}
}
\end{table}

The input layer is of the size $784$, hidden layers H\textsubscript{1} to H\textsubscript{5} are activated by ReLU, H\textsubscript{6} is activated by linear function to represent the logits, and, finally, the output layer is a Softmax layer to normalize the logits and produce the probabilities. Output layer only enabled during training step.

\subsection{Experiment problem}
As highlighted in Section~\ref{background}, trained DNNs have a variety of applications. These include feature visualization \citep{fischetti2018deep}, DNNs verification problem \citep{dutta2017output,cheng2017maximum,bunel2018unified,anderson2020strong,tjeng2018evaluating,bunel2020branch}, transforming a NN to another \citep{kumar2019equivalent}, reachability analysis of NNs \citep{lomuscio2017approach}, counting and bounding linear regions to evaluate the expressiveness of NNs \citep{serra2018bounding,serra2020empirical}, identifying critical neurons in the network \citep{elaraby2020identifying}, and finding adversarial examples \citep{anderson2020strong,cheng2017maximum,fischetti2018deep,tsay2021partition}.

One of the most intriguing applications of MIP technology in DNNs is creating adversarial examples, a concept central to adversarial machine learning \citep{szegedy2014intriguing}. This involves slightly modifying a given input to an FF-DNN to produce an incorrect output, exposing potential weaknesses in the network. Constructing these optimized adversarial examples is critical for understanding and improving the robustness of DNNs. Following the same setting as \cite{fischetti2018deep}, consider an input image $\Tilde{x}^{0}$ that is correctly classified by a network as a specific digit $\Tilde{d}$. The goal is to generate a similar image $x^{0}$ that is mistakenly classified as a different digit $d\ne\Tilde{d}$.

We introduce an intentional misclassification of the actual (incorrect) digit $d$, which we aim to achieve by defining $d=(\widetilde{d}+5)\mod10$. For instance, we stipulate that a “0” should be classified as a “5” and a “6” as a “1”. To enforce this condition, a constraint is imposed in the final layer of the network, ensuring that the activation corresponding to the incorrect digit is at least $20\%$ greater than the activation for any other digits. This requirement is incorporated by adding a specific linear inequality, denoted as \eqref{eq:3.1}, to each IP formulation employed in the experiments.

\begin{equation}\label{eq:3.1}    x_{d+1}^{L}\geq1.2x_{j}^{L}\hspace{1cm}\forall{j\in\{0,\dots,9\}\setminus\{d\}}
\end{equation}
To minimize the $L_1$-norm between $x^{0}$ and $\Tilde{x}^{0}$, the objective function $\sum_{j=1}^{n^{0}}d_{j}$ is used. Here, $d_j$'s are additional continuous variables introduced to ensure the set of linear inequalities \eqref{eq:3.2}:
\begin{equation}\label{eq:3.2}
    -d_{j}\leq x^{0}_{j}-\Tilde{x}^{0}_{j}\leq d_{j},\hspace{5mm}d_{j}\geq0,\hspace{5mm}\forall{j=1,\dots,n^{0}}
\end{equation}
The constraints set \eqref{eq:3.2} is also added to each MIP formulation used in the experiments.

\subsection{Computational performance}
In this section, we examine the empirical performance of a state-of-the-art MIP solver in generating adversarial examples for moderately sized FF-DNNs, as detailed in Table \ref{tab:set}. All experiments use the Gurobi v9.0.1 solver, executed with 4 threads on a machine equipped with 8 GB of RAM and an Intel(R) Core i7-7660U CPU 2.50GHz. A time limit of 30 minutes (1800 s) is set for each run, and the default settings of Gurobi are retained. We consider $\Tilde{d}=4$ for set of inequalities \eqref{eq:3.1} and \eqref{eq:3.2}.

Table \ref{tab:result} represents the results of implementing the discussed formulations in Section 3 on the mentioned problem for trained Networks 1 and 2, whose architectures are discussed in Table \ref{tab:set}. We note that the proposed method for Big-$M$ by R. Anderson requires an exponential number of constraints \citep{anderson2020strong}. In our experiment, it took more than an hour for Gurobi to build the model, and it could not build it even for Network 1. Therefore, we did not experiment with that method. For the remaining approaches, first, we solved the corresponding basic formulations, and then, we added the proposed valid inequality.

\begin{table}
\resizebox{\linewidth}{!}{%
\centering
\caption{Experimental results of implementing trained ReLU FF-DNNs}
\label{tab:result}
\begin{tabular}{|c|c|cc|c|c|ccccc|ccccc|}
\cline{2-16}
\multicolumn{1}{c|}{\multirow{2}{*}{}}        & \multirow{2}{*}{\textbf{Method}} & \multicolumn{2}{c|}{\textbf{Variable}} & \multirow{2}{*}{\textbf{Constraint}}                   & \multirow{2}{*}{\textbf{LP}} & \multicolumn{5}{c|}{\textbf{Basic}}                                                & \multicolumn{5}{c|}{\textbf{Valid Inequality}}                                      \\ 
\cline{3-4}\cline{7-16}
\multicolumn{1}{c|}{}                         &                                  & \textbf{Continuous} & \textbf{Binary}  &                                                        &                              & \textbf{Time (s)} & \textbf{Cut} & \textbf{Node} & \textbf{gap} & \textbf{Optimal} & \textbf{Time (s)} & \textbf{Cut} & \textbf{Node} & \textbf{gap} & \textbf{Optimal}  \\ 
\hline
\multirow{5}{*}{\textbf{Network 1}} & \textbf{Big-$M$-B\tablefootnote{Big-$M$ method by R.Bunel's proposed bounds}}                 & $1648$              & $70$             & \begin{tabular}[c]{@{}c@{}}$1797$\\$1897$\end{tabular} & $0.0$                        & $3.93$            & $9$          & $1231$        & $0.00\%$     & $44.10$           & $17$              & $87$         & $1576$        & $0.00\%$     & $44.10$            \\
& \textbf{Big-$M$-C\tablefootnote{Big-$M$ method by C.Cheng's proposed bounds}}                 & $1648$              & $70$             & \begin{tabular}[c]{@{}c@{}}$1797$\\$1897$\end{tabular} & $0.0$                        & $17.5$            & $14$         & $8447$        & $0.00\%$     & $44.10$           & $22.1$            & $7$          & $8441$        & $0.00\%$     & $44.10$            \\
& \textbf{Extended}                & $1718$              & $70$             & \begin{tabular}[c]{@{}c@{}}$1797$\\$1897$\end{tabular} & $0.0$                        & $5.68$            & $7$          & $1921$        & $0.00\%$     & $44.10$           & $5.7$             & $3$          & $2178$        & $0.00\%$     & $44.10$            \\
& \textbf{Disjunctive}             & $1788$              & $70$             & $2147$                                                 & $0.0$                        & $3.93$            & $4$          & $1118$        & $0.00\%$     & $44.10$           & -                 & -            & -             & -            & -                 \\
& \textbf{Big-$M$-A\tablefootnote{R.Anderson's proposed method}}                 & -                   & -                & -                                                      & -                            & -                 & -            & -             & -            & -                & -                 & -            & -             & -            & -                 \\ 
\hline
\multirow{5}{*}{\textbf{Network 2}} & \textbf{Big-$M$-B}                 & $2074$              & $496$            & \begin{tabular}[c]{@{}c@{}}$3075$\\$3555$\end{tabular} & $0.0$                        & $1800$            & $24$         & $6880$        & -            & -                & $1800$            & $1$          & $6272$        & $100\%$      & $89.49$           \\
& \textbf{Big-$M$-C}                 & $2074$              & $496$            & \begin{tabular}[c]{@{}c@{}}$3075$\\$3555$\end{tabular} & $0.0$                        & $1800$            & $48$         & $6822$        & -            & -                & $1800$            & $2$          & $5637$        & -            & -                 \\
& \textbf{Extended}                & $2570$              & $496$            & \begin{tabular}[c]{@{}c@{}}$3075$\\$3555$\end{tabular} & $0.0$                        & $1800$            & $23$         & $13368$       & -            & -                & $1800$            & $1$          & $9993$        & -            & -                 \\
& \textbf{Disjunctive}             & $3066$              & $496$            & \begin{tabular}[c]{@{}c@{}}$5555$\\$6035$\end{tabular} & $0.0$                        & $1800$            & $5$          & $9803$        & $100\%$      & $73.36$          & $1800$            & $10$         & $6543$        & $100\%$      & $81.23$           \\
& \textbf{Big-$M$-A}                 & -                   & -                & -                                                      & -                            & -                 & -            & -             & -            & -                & -                 & -            & -             & -            & -                 \\
\hline
\end{tabular}
}
\end{table}

Table \ref{tab:result} denotes the number of variables (continuous and binary) and constraints (basic formulation and after adding valid inequality), the optimal solution of the LP-relaxation of each approach, and, for the MIP, time, number of cuts that Gurobi added, number of nodes Gurobi explored, the gap between best bound and the optimal value, and the optimal value. As can be seen, Gurobi performed well on Network 1 (narrower architecture) and could not find the bound for the broader network since, by going deeper and wider in the architecture, the number of constraints and variables grows exponentially, solving a problem to optimality is computationally expensive. Among the methods in Network 1, the extended formulation performs better than the others since the valid inequalities are proposed for this formulation inherently. Also, disjunctive formulation had a good performance since the number of variables and constraints for this approach is higher than the others, and it has the same results in a lower time.

\section{Discussions and Conclusions}\label{conclusions}
Our investigation has focused on the MIP formulation of DNN incorporating ReLUs and maximum/average pooling. This marks an initial stride in leveraging discrete optimization as a fundamental tool for analyzing NNs. Benefiting from the structure of the ReLU activation function, given that weights and biases of the network are available, the entire trained NN model is interpreted as a piecewise linear function. Hence, introducing some binary and/or auxiliary variables can formulate a trained DNN model as MIP. This technique helps us to formally investigate the behavior of the NNs model, especially in hidden layers, which makes it not a black box. As seen from the literature, this technique is of interest to researchers from various points of view. We discussed the different applications of trained ReLU networks MIP formulation. While these formulations are not well-suited for training, as they become bilinear in this context, their utility lies in constructing optimized input examples for a pre-trained NN. In this vein, we have discussed their application in generating adversarial examples, contributing to the Machine Learning community's understanding and exploration of such techniques. 

We trained two not-to-shallow NNs on handwritten digit classification using the MNIST dataset, such that Network 2 is more expansive than Network 1. As a comparative study, we established each existing formulation for ReLU on these networks. Gurobi, with default parameters, efficiently found an optimal solution for each formulation for Network 1. However, Network 2 could not find a best bound for MIP. From the results of Network 1, we can see that extended and disjunctive formulation was fast and reliable. However, when the network gets wider, \emph{i.e.}, the model contains more variables and constraints, and the solver could not find a proper bound on the best solution in $30$ minutes. The configuration of the machine in which we implement the model is influential in finding the optimal solution. One can establish the model on a more robust system and obtain optimal solutions for broader and deeper networks. Also, we could not implement the Big-M formulation proposed by Anderson et al. since it produces exponentially many constraints. In our experiments, building up a model by Gurobi for Anderson's formulation took more than an hour for Network 1.

Other than the proposed formulations for ReLU DNNs, a new line of research has been emerged for training discrete DNNs by the MIP technique. We discussed the formulations of binary and binarized DNNs in Section 3. Since their purposes and formulations differ from those of trained NNs, it is not reasonable to compare them. Therefore, the interested reader is referred to those papers for further studies.

Another application of trained DNNs is recently emerged in the management science and operations research community. The trained ReLU NN MIP formulation is used in the prescriptive analysis of the data-driven optimization models.

\bibliographystyle{unsrtnat}
\bibliography{references}  






\end{document}